\documentclass[10pt, a4paper]{article}

\usepackage{booktabs}
\usepackage{multirow}

\usepackage[final]{lrec2026} 

\title{ManufactuBERT: Efficient Continual Pretraining for Manufacturing}

\name{Robin Armingaud, Romaric Besançon} 

\address{Universit\'e Paris-Saclay, CEA, List, F-91120, Palaiseau, France \\
         \{firstname.lastname\}@cea.fr\\}

\abstract{
While large general-purpose Transformer-based encoders excel at general language understanding, their performance diminishes in specialized domains like manufacturing due to a lack of exposure to domain-specific terminology and semantics. In this paper, we address this gap by introducing ManufactuBERT, a RoBERTa model continually pretrained on a large-scale corpus curated for the manufacturing domain. We present a comprehensive data processing pipeline to create this corpus from web data, involving an initial domain-specific filtering step followed by a multi-stage deduplication process that removes redundancies. Our experiments show that ManufactuBERT establishes a new state-of-the-art on a range of manufacturing-related NLP tasks, outperforming strong specialized baselines. More importantly, we demonstrate that training on our carefully deduplicated corpus significantly accelerates convergence, leading to a 33\% reduction in training time and computational cost compared to training on the non-deduplicated dataset. The proposed pipeline offers a reproducible example for developing high-performing encoders in other specialized domains. We will release our model and curated corpus at \url{https://huggingface.co/cea-list-ia}
 \\ \newline  }

\begin{document}

\maketitleabstract

\section{Introduction}

The digitalization of the manufacturing sector has led to an explosion of textual data, making Natural Language Processing (NLP) methods especially efficient for tasks such as automated anomaly detection, technical report completion and knowledge extraction (\citealp{MAY2022184}; \citealp{NLPINMANUFACTURING}; \citealp{li2024largelanguagemodelsmanufacturing}). Transformer-based encoders, from BERT\cite{devlin-etal-2019-bert} to its successors RoBERTa \cite{liu2019robertarobustlyoptimizedbert}, DeBERTa \cite{he2021debertav3} and more recently NeoBERT \cite{breton2025neobertnextgenerationbert} and ModernBERT \cite{modernbert}, have achieved near-human-level performance on a wide range of language understanding benchmarks, including Named Entity Recognition (NER) and Relation Extraction (RE). However, the effectiveness of these general-purpose models is often limited when applied to specialized domains. Industrial and manufacturing texts exhibit unique linguistic properties, including specialized terminology, a high frequency of acronyms and context-dependent meanings that differ from common usage. The standard approach to bridge this gap is domain adaptation \cite{gururangan-etal-2020-dont}, which involves continuing the pretraining of a model on a large, domain-specific corpus. While effective, this process is computationally expensive, requiring significant resources and contributing to a larger carbon footprint.
\begin{figure}[!ht]
\begin{center}
\includegraphics[width=\columnwidth]{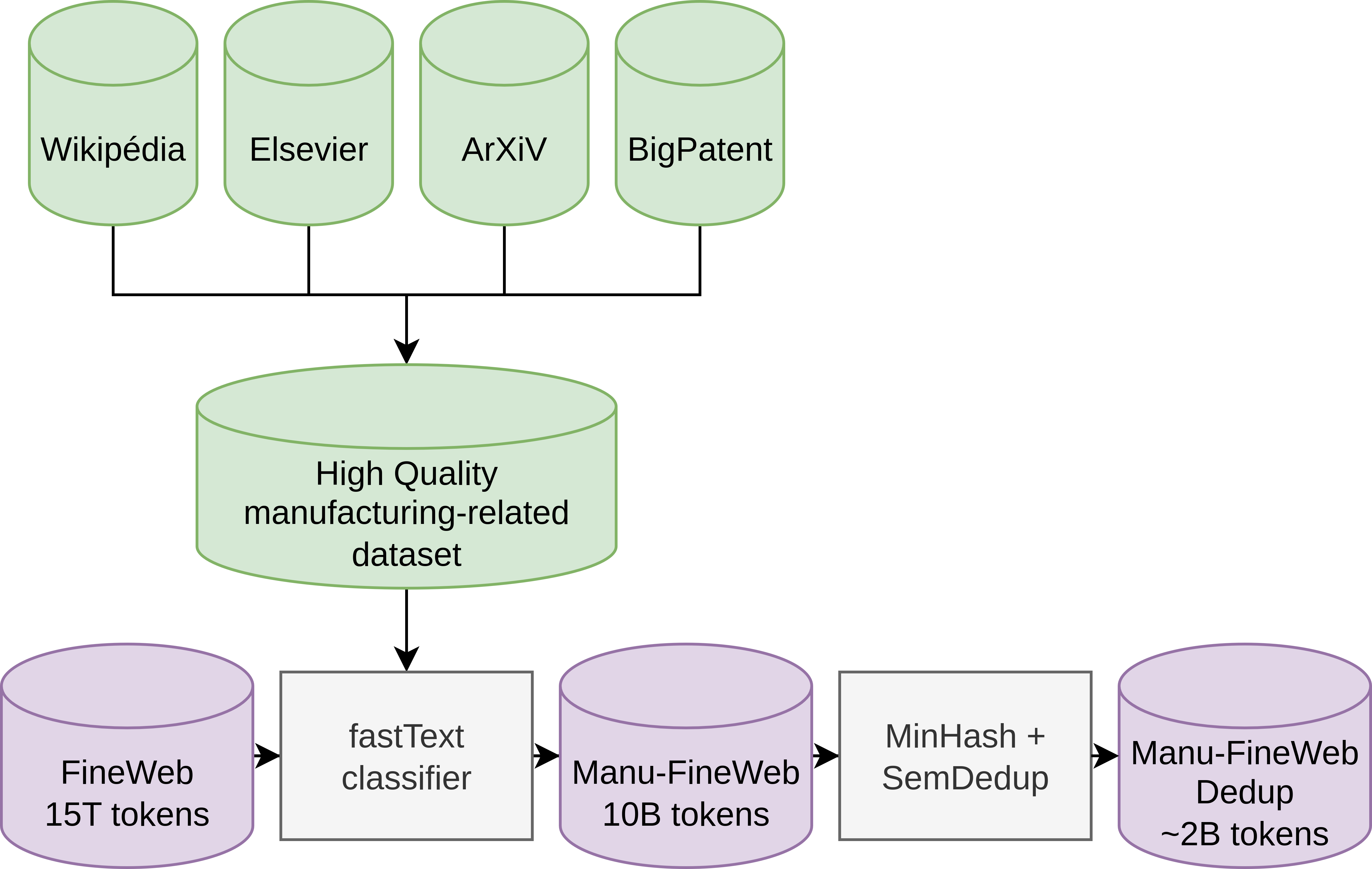}
\caption{Workflow of the data filtering and deduplication steps used to create the ManufactuBERT pretraining corpus.}
\label{fig:data_pipeline}
\end{center}
\end{figure}

This work addresses the challenge of efficiently adapting language models to the manufacturing domain. We introduce a pipeline for creating a high-quality, domain-specific corpus and a new pretrained language model, ManufactuBERT. We propose the following contributions : 

\begin{itemize}
    \item We introduce ManufactuBERT, a new Pretrained Language Model (PLM) based on RoBERTa, pretrained on a large-scale, curated corpus of manufacturing-related texts.
    \item We propose an efficient adaptation pipeline for encoder-based language models, designed to reduce the data and energy footprint of domain specialization : we construct a manufacturing-oriented pretraining corpus by filtering the FineWeb dataset \cite{penedo2024the} using a fastText-based domain classifier \cite{joulin-etal-2017-bag}, followed by deduplication using SemDeDup \cite{abbas2023semdedup}. This approach yields a compact yet representative dataset that accelerates training convergence and reduces storage requirements.
    \item We conduct a comprehensive evaluation of ManufactuBERT against strong baselines on a suite of tasks relevant to the manufacturing domain, including the FabNER benchmark \cite{DBLP:journals/jim/KumarS22}. Our results show that ManufactuBERT establishes a new state-of-the-art on several of these tasks.
\end{itemize}

\section{Related Work}

\paragraph{Transformer models}
Since their introduction, encoder-only Transformer models like BERT \cite{devlin-etal-2019-bert}, have become foundational in NLP. This architecture has inspired a family of successors. For instance, RoBERTa \cite{liu2019robertarobustlyoptimizedbert} revisited BERT's pretraining strategy, showing that removing the next-sentence prediction objective, increasing batch sizes, using dynamic masking and training on more data yields stronger performance. DeBERTa / DeBERTaV3 (\citealp{he2021deberta}; \citealp{he2021debertav3}) improves upon previous encoders by introducing disentangled attention, enhanced position encodings and more efficient hyperparameters.

More recently, models such as ModernBERT and NeoBERT have started to incorporate more training data and architectural concepts from modern Large Language Models (LLMs) into the BERT framework. Despite these advances, a common limitation persists: these models are trained on general-domain corpora, which restricts their applicability in specialized fields without further adaptation to learn domain-specific vocabularies and semantic nuances.

\paragraph{Domain Adaptation}
One of the standard approaches to adapt a general-purpose PLM to a specialized domain is continued pretraining on in-domain unlabeled text, referred to as domain-adaptive pretraining. This involves continuing the Masked Language Modeling (MLM) pretraining objective. \citet{gururangan-etal-2020-dont} provided strong evidence that domain-adaptive pretraining yields performance gains on downstream tasks.

This principle has led to the development of numerous domain-specific models. Notable examples include BioBERT \cite{Lee_2019} which adapts BERT by further pretraining on large corpora from PubMed abstracts and PMC full texts or FinBERT \cite{ijcai2020p622} which adapt BERT to financial domain texts.
Some models, like SciBERT \cite{beltagy2019scibertpretrainedlanguagemodel}, are trained from scratch on scientific publications, allowing for the creation of a new and domain-aligned vocabulary. 

Closer to our work, MatSciBERT \cite{gupta2021matscibert} specializes in materials science by continuing the pretraining of SciBERT on a targeted corpus of scientific articles. 

While alternative adaptation methods exist, such as vocabulary expansion techniques like AVocaDo \cite{hong-etal-2021-avocado}, they have shown limited efficacy with more robust encoders like RoBERTa. Furthermore, \citet{kim-etal-2024-seed} suggests that these methods may offer limited benefits in highly technical domains like materials science , without additional tuning.

\paragraph{Data Selection for Pretraining}
The high cost of pretraining is linked to the massive scale of the datasets used, which are often derived from web crawls like Common Crawl such as C4 \cite{2019t5} or more recently FineWeb \cite{abbas2023semdedup} and RefinedWeb \cite{refinedweb}. While these corpora apply extensive filtering, they still contain significant redundancy. To address this issue, recent work has focused on optimizing data selection to improve training efficiency and model performance. This has led to the development of sophisticated deduplication algorithms that go beyond simple lexical matching such as Minhash \cite{broder1997resemblance}. Approaches like SemDeDup \cite{abbas2023semdedup}, Density Based Pruning \cite{kamal2024density} and D4 \cite{tirumala2023d4improvingllmpretraining} use embeddings to identify and remove similar documents and increase data diversity. While these methods have primarily been applied to image models or LLMs, we employ SemDeDup to clean our pretraining corpus, and to the best of our knowledge, we are the first to use this approach in the context of MLM.

\section{Methodology}

\subsection{Domain-specific pretraining dataset construction}

While many domain-specific encoders, such as SciBERT, are pretrained on curated and homogeneous corpora like scientific literature, which leads to excellent performance on academic benchmarks built on the same corpora such as SciERC or SciCite (\citealp{luan-etal-2018-multi}; \citealp{cohan-etal-2019-structural}), this approach can create a domain mismatch with the linguistic diversity found in real-world industrial applications. 
To pretrain ManufactuBERT, we instead rely on a specific selection and curation process from a large-scale, web-based diverse corpus. This process, illustrated in Figure~\ref{fig:data_pipeline} is composed of two steps: filtering and deduplication.

\begin{table*}[!ht]
\centering
\begin{tabular}{p{3cm}p{9cm}r}
\hline
\textbf{Data Source} & \textbf{Selection Criteria} & \textbf{Documents} \\
\hline
\textbf{Elsevier} & Abstracts from manufacturing-related journals retrieved via the Elsevier API, based on the SciMago journal rankings for the "Industrial and Manufacturing Engineering" category. & 27\,943 \\
\textbf{ArXiv} & Abstracts from the cond-mat, physics and eess categories containing keywords such as \textit{manufacturing}, \textit{3D printing}, or \textit{industrial process}, following a similar methodology to \citet{DBLP:journals/jim/KumarS22}. & 2\,042 \\
\textbf{Wikipedia} & Articles from categories including "Manufacturing", "Engineering" and "Industrial processes".  & 5\,907 \\
\textbf{BigPatent} \citep{sharma-etal-2019-bigpatent} &  Patent descriptions containing the keyword "manufacturing". & 26\,428 \\
\hline
\end{tabular}
\caption{Training data for the manufacturing-domain classifier used to filter FineWeb.}
\label{tab:classifier_data}
\end{table*}

\subsubsection{Data Filtering}
We select as base corpus the recent and high-quality FineWeb dataset \cite{penedo2024the}, derived from Common Crawl and comprising approximately 15 trillion tokens. 
We then filter this dataset using a classifier trained to identify documents that are relevant to the manufacturing domain. We use a FastText classifier \cite{joulin-etal-2017-bag}, chosen for its efficiency and effectiveness in tasks such as domain or language identification. The classifier is trained on a curated dataset with positive examples selected from different relevant sources and negative examples randomly selected from FineWeb, with a negative-to-positive ratio of 10:1. The sources and criteria used for the positive examples are summarized in Table~\ref{tab:classifier_data}. Note that this dataset is too small to support effective pretraining and we use it only to train the classifier.

We filter FineWeb using the trained classifier and adjust the prediction threshold to obtain a corpus of approximately 10 billion tokens (around 21 million documents), comparable in scale to the corpora used by \citet{gururangan-etal-2020-dont}.

\subsubsection{Data Deduplication}

Since FineWeb is composed of multiple individually deduplicated Common Crawl snapshots, it still contains a substantial amount of residual duplicates. To further improve the quality and diversity of our pretraining corpus, we apply SemDeDup \cite{abbas2023semdedup}, a deduplication algorithm designed to identify and remove semantically redundant documents. Following the authors' recommendations, we first perform MinHash deduplication with 20 buckets and 20 signatures per bucket to eliminate exact and near-lexical duplicates.
The SemDeDup process then consists of three main steps:
\begin{enumerate}
    
\item Computing a semantic vector representation for each document in the corpus. For this task, we employ the all-MiniLM-L6-v2 sentence encoder, a computationally efficient and widely-used model for semantic similarity tasks. To handle documents that exceed the model's maximum input length, we segment each document into chunks of 512 tokens. Each chunk is independently encoded using the SentenceTransformers library\footnote{\url{https://www.sbert.net/}}, and the resulting chunk embeddings are then averaged to produce a single vector representing the entire document.

\item Partitioning the embeddings into $n$ clusters using K-means 

\item Within each cluster, pruning any document that is closer to another document than a specified distance threshold $\tau$. 
\end{enumerate}

For our final corpus, we set the number of clusters $n=1000$ and the distance threshold $\tau=0.15$. This combined MinHash and SemDeDup process effectively removes approximately 80\% of the documents from our initial filtered manufacturing dataset.
We implement this pipeline using the Datatrove library \cite{penedo2024datatrove}, originally released with the FineWeb dataset, along with a custom implementation of SemDeDup. The resulting corpus of approximately 4.5 million documents is highly aligned with the target domain while maintaining strong general-domain performance, as demonstrated in Section~\ref{sec:experiments}. Moreover, the proposed pipeline is easily reproducible for other domains, provided that a domain classifier can be trained.

\subsection{Masked Language Modeling Pretraining}

We perform continued pretraining, initializing our model with the publicly available RoBERTa-base checkpoint. The training follows the standard Masked Language Modeling objective, omitting the Next Sentence Prediction task, which is consistent with the methodology of RoBERTa.

We adopt the hyperparameters from \citet{gururangan-etal-2020-dont}: a batch size of 16 with 16 gradient accumulation steps (for an effective batch size of 256 per GPU and 2048 in total), a masking probability of 0.15, a weight decay of 0.1 and a maximum learning rate of $5 \times 10^{-4}$ with a linear scheduler and 6\% warmup steps. The model is checkpointed every 500 steps. We make one modification: we extend the training schedule from 12,500 to 17,500 steps, as our preliminary experiments showed that the model had not yet converged at 12,500 steps. This pretraining phase was executed on a node of 8 NVIDIA V100-32GB GPUs and required approximately 51 hours of computation.
We pretrain two models: ManufactuBERT, using the non-deduplicated dataset and ManufactuBERT\textsubscript{D}, using the deduplicated corpus. The latter converges faster and achieves higher performance.

\section{Experiments}\label{sec:experiments}
\subsection{Datasets}
To evaluate our model, we select a range of datasets related to the manufacturing domain. Due to the limited availability of annotated resources in this area, we also include datasets from adjacent technical domains to assess our model's generalization across diverse tasks. Our selection criteria include data accessibility, task diversity and domain proximity. The chosen datasets span three core NLP tasks: Sentence Classification (SC), Named Entity Recognition (NER) and Relation Extraction (RE).

We evaluate on the following benchmarks:
\begin{itemize}
    \item \textbf{FabNER} \cite{DBLP:journals/jim/KumarS22} : A manufacturing-domain corpus for NER, consisting of approximately 14k scientific abstracts.

\item \textbf{Materials Synthesis} \cite{mysore-etal-2019-materials} : A dataset of 230 abstracts describing materials synthesis procedures. Its labeled graphs enable both NER and RE evaluation.

\item \textbf{SOFC} \cite{friedrich-etal-2020-sofc} : A corpus of 45 expert-annotated research articles on solid oxide fuel cells (SOFCs). The dataset defines four NER entity types,Material, Experiment, Value and Device and an extended slot-filling version (SOFC-Slot) with 16 fine-grained entity types. It also includes sentence-level annotations for SC, distinguishing SOFC-related and non-related content.

\item \textbf{MatScholar} \cite{doi:10.1021/acs.jcim.9b00470} : A hand-annotated corpus of 800 materials science abstracts for NER.

\item \textbf{Big Patent} \cite{sharma-etal-2019-bigpatent} : A large-scale dataset of U.S. patents categorized into nine classes. We randomly sample 1,000 documents per class for training, 200 for validation and 200 for testing, and perform document classification based on the abstracts.

\item \textbf{ChemdNER} \cite{Krallinger2015} : A dataset of 10,000 PubMed abstracts annotated for chemical entities. While originating from the biomedical domain, this corpus serves as a benchmark to evaluate our model's ability to identify chemical terms. It is useful for advanced manufacturing sectors, such as pharmaceutical manufacturing.
\end{itemize}

While our focus is on domain-specific performance, it is also crucial to assess whether the adaptation process has degraded the model's language understanding capabilities in general-domain. To this end, we also evaluate ManufactuBERT on the General Language Understanding Evaluation (GLUE) benchmark \citep{wang2019glue}. This evaluation serves to quantify any potential loss in general-domain performance and allows for a direct comparison with other specialized models, MatSciBERT and SciBERT. Following standard evaluation practices, we exclude the WNLI task from the benchmark.

\subsection{Experimental Setup}\label{sec:hyperparameters_finetuning}

For all manufacturing-related downstream tasks, we fine-tune each model for a maximum of 20 epochs, reporting results across 5 different random seeds. For each run, we select the checkpoint that achieves the highest performance on the development set for final evaluation. 

For the GLUE benchmark, we fine-tune the models for 10 epochs on each dataset. For all experiments, we report the mean and standard deviation of the results. Our implementation is based on the HuggingFace Transformers library \citep{wolf-etal-2020-transformers}. For GLUE, we use the official text classification script provided in the Transformers repository\footnote{\url{https://github.com/huggingface/transformers/blob/main/examples/pytorch/text-classification/run_glue.py}}.

We use a fixed learning rate of $2 \times 10^{-5}$, a batch size of 16, a weight decay of 0.1, the AdamW optimizer \cite{loshchilov2019decoupledweightdecayregularization} and a linear learning rate scheduler similar to \citet{gururangan-etal-2020-dont} experiments. 

All experiments are conducted on a single NVIDIA A100 or H100 GPU.

\subsection{Results}

\begin{table*}[ht]
\centering
\resizebox{\linewidth}{!}{
\begin{tabular}{lcccccccccc}
\toprule
\multirow{2}{*}{\textbf{Model}}    & \multicolumn{1}{c}{\textbf{Materials Synthesis}}                         & \multicolumn{1}{c}{\textbf{FabNER}}    & \multicolumn{2}{c}{\textbf{SOFC}}     &   \multicolumn{1}{c}{\textbf{MatScholar}}   &   \multicolumn{1}{c}{\textbf{ChemdNER}}    &  \multicolumn{1}{c}{} \\  \cmidrule(r){2-2} \cmidrule(r){3-3} \cmidrule(r){4-5} \cmidrule(r){6-6} \cmidrule(r){7-7}
                           & NER & NER & NER & NER SLOT & NER & NER & \textbf{Avg.} \\
\midrule
ModernBERT & $69.60_{\pm 1.71}$ & $78.76_{\pm 0.68}$ & $74.62_{\pm 1.43}$ & $53.44_{\pm 1.80}$ & $80.14_{\pm 0.46}$ & $89.12_{\pm 0.25}$ & $74.28$ \\
NeoBERT & $73.00_{\pm 1.53}$ & $82.36_{\pm 0.27}$ & $74.14_{\pm 0.89}$ & $57.90_{\pm 4.08}$ & $81.56_{\pm 0.57}$ & $90.40_{\pm 0.23}$ & $76.56$ \\
RoBERTa & $73.12_{\pm 0.35}$ & $82.48_{\pm 0.33}$ & $82.54_{\pm 0.88}$ & $69.52_{\pm 1.52}$ & $84.04_{\pm 0.18}$ & $90.50_{\pm 0.20}$ & $80.37$ \\
SciBERT & $77.72_{\pm 0.41}$ & $83.60_{\pm 0.28}$ & $80.34_{\pm 0.48}$ & $69.10_{\pm 0.68}$ & $84.52_{\pm 0.13}$ & $91.80_{\pm 0.10}$ & $81.18$ \\
DeBERTaV3 & $73.92_{\pm 0.54}$ & $\mathbf{84.62}_{\pm 0.16}$ & $82.86_{\pm 0.77}$ & $70.68_{\pm 0.98}$ & $85.04_{\pm 0.49}$ & $91.74_{\pm 0.17}$ & $81.48$ \\
MatSciBERT & $\mathbf{76.50}_{\pm 0.87}$ & $83.88_{\pm 0.19}$ & $82.10_{\pm 0.57}$ & $72.60_{\pm 1.34}$ & $85.88_{\pm 0.32}$ & $\mathbf{92.00}_{\pm 0.19}$ & $82.16$ \\
\textbf{ManufactuBERT} & $75.20_{\pm 0.60}$ & $83.88_{\pm 0.18}$ & $83.96_{\pm 0.76}$ & $73.64_{\pm 0.85}$ & $86.06_{\pm 0.32}$ & $91.94_{\pm 0.21}$ & $82.45$ \\
\textbf{ManufactuBERT\textsubscript{D}} & $75.04_{\pm 0.22}$ & $84.00_{\pm 0.24}$ & $\mathbf{84.40}_{\pm 0.34}$ & $\mathbf{73.68}_{\pm 1.05}$ & $\mathbf{86.76}_{\pm 0.34}$ & $91.92_{\pm 0.16}$ & $\mathbf{82.63}$ \\
\bottomrule
\end{tabular}
}
\caption{{\textbf{Manufacturing Domain NER}: Micro-averaged F1 scores across six manufacturing-related NER tasks. ManufactuBERT\textsubscript{D} denotes the model trained on the deduplicated dataset. Best results per column are shown in \textbf{bold}.}}
\label{tab:manubenchmark_ner}
\end{table*}

\begin{table*}[ht!]
\centering
\begin{tabular}{lcccccccccc}
\toprule
\multirow{2}{*}{\textbf{Model}}    & \multicolumn{1}{c}{\textbf{Materials Synthesis}}                            & \multicolumn{1}{c}{\textbf{SOFC}}     &  \multicolumn{1}{c}{\textbf{Big Patent}}   &  \multicolumn{1}{c}{} \\  \cmidrule(r){2-2} \cmidrule(r){3-3} \cmidrule(r){4-4}
                           & RE & SC & SC & \textbf{Avg.} \\
\midrule
ModernBERT & $95.10_{\pm 0.22}$ & $94.36_{\pm 0.26}$ & $63.12_{\pm 0.24}$ & $84.19$ \\
NeoBERT & $94.36_{\pm 0.37}$ & $94.46_{\pm 0.15}$ & $64.58_{\pm 0.63}$ & $84.47$ \\
RoBERTa & $94.32_{\pm 0.22}$ & $94.32_{\pm 0.15}$ & $63.58_{\pm 0.68}$ & $84.07$ \\
SciBERT & $95.34_{\pm 0.17}$ & $\mathbf{94.76}_{\pm 0.18}$ & $63.98_{\pm 0.46}$ & $84.69$ \\
DeBERTaV3 & $\mathbf{95.80}_{\pm 0.20}$ & $94.40_{\pm 0.16}$ & $64.46_{\pm 0.54}$ & $84.89$ \\
MatSciBERT & $95.48_{\pm 0.26}$ & $94.72_{\pm 0.34}$ & $63.80_{\pm 0.58}$ & $84.67$ \\
\textbf{ManufactuBERT} & $94.60_{\pm 0.22}$ & $94.46_{\pm 0.15}$ & $65.50_{\pm 0.35}$ & $84.85$ \\
\textbf{ManufactuBERT\textsubscript{D}} & $94.62_{\pm 0.11}$ & $94.68_{\pm 0.13}$ & $\mathbf{65.80}_{\pm 0.31}$ & $\mathbf{85.03}$ \\
\bottomrule
\end{tabular}
\caption{{\textbf{Manufacturing Domain RE and SC}: Micro-averaged F1 scores across three manufacturing-related classification tasks. ManufactuBERT\textsubscript{D} denotes the model trained on the deduplicated dataset. Best results per column are shown in \textbf{bold}.}}
\label{tab:manubenchmark_classif}
\end{table*}

\subsubsection{Manufacturing-Related Tasks}

We conduct a comprehensive evaluation of our models on nine manufacturing-related tasks. The models are benchmarked against a suite of strong baselines: 
\begin{itemize}
    \item general-purpose encoders: RoBERTa \cite{liu2019robertarobustlyoptimizedbert}, DeBERTaV3 \cite{he2021debertav3}, NeoBERT \cite{breton2025neobertnextgenerationbert} and ModernBERT \cite{modernbert} 
    \item domain-specific models : SciBERT \cite{beltagy2019scibertpretrainedlanguagemodel} and MatSciBERT \cite{gupta2021matscibert}. 
\end{itemize}
To ensure a fair comparison, we use the base version for all models. The detailed results, reported as micro F1 scores, are presented in Table \ref{tab:manubenchmark_classif} and Table \ref{tab:manubenchmark_ner}.

ManuBERT \cite{kumar2023manubert}, a non-peer-reviewed model pretrained for the manufacturing domain, was excluded from our main benchmark as its reported results could not be reproduced using the publicly available checkpoint\footnote{\url{https://huggingface.co/akumar33/ManuBERT}}
. Its performance on FabNER is shown in Table~\ref{tab:manubert_results} for reference.

\begin{table}[h]
\centering
\begin{tabular}{lcc}
\toprule
\textbf{Model} & \textbf{Micro F1} & \textbf{Macro F1} \\
\midrule
ManuBERT & $82.60{\pm 0.21}$ & $76.52_{\pm 0.67}$ \\
RoBERTa & $82.48{\pm 0.33}$ & $77.30_{\pm 0.89}$ \\
ManufactuBERT\textsubscript{D} & $\mathbf{84.00_{\pm 0.24}}$ & $\mathbf{78.98_{\pm 0.37}}$ \\
\bottomrule
\end{tabular}
\caption{Comparison of ManuBERT, RoBERTa and ManufactuBERT\textsubscript{D} on FabNER. ManuBERT performs similarly to RoBERTa.}
\label{tab:manubert_results}
\end{table}

We group classification-related tasks (Sentence Classification and Relation Extraction) in Table~\ref{tab:manubenchmark_classif} and NER tasks in Table~\ref{tab:manubenchmark_ner}. The results demonstrate the effectiveness of our domain adaptation strategy. Our primary model, ManufactuBERT\textsubscript{D}, achieves the highest average F1 score across both groups of tasks, outperforming the strongest domain-specific baseline, MatSciBERT, and the original RoBERTa model. This confirms that our pretraining corpus effectively captures knowledge relevant to the manufacturing domain. ManufactuBERT\textsubscript{D} establishes new state-of-the-art performance on four of the nine evaluated tasks.

Moreover, the model trained on the semantically deduplicated corpus, ManufactuBERT\textsubscript{D}, consistently outperforms its counterpart trained on the non-deduplicated data ManufactuBERT. This result underscores the value of using SemDeDup for MLM-based domain adaptation, as it leads to better downstream performance even with a smaller pretraining dataset.

\begin{table*}[ht!]
\centering
\begin{tabular}{lccccccccc}
\toprule
\textbf{Model} & \textbf{CoLA} & \textbf{MNLI} & \textbf{MRPC} & \textbf{SST-2}  & \textbf{QNLI} & \textbf{RTE} & \textbf{STS-B}  & \textbf{QQP} & \textbf{Avg.} \\
\midrule
\multicolumn{10}{l}{\textit{General Domain Model}}\\
RoBERTa & $63.6$ & $87.6$ & $90.2$ & $94.8$ & $92.8$ & $78.7$ & $91.2$ & $91.9$ & $86.35$ \\
\midrule
\multicolumn{10}{l}{\textit{Specialized Models}} \\
MatSciBERT & $34.27$ & $81.03$ & $84.40$ & $88.91$ & $88.23$ & $62.58$ & $85.83$ & $90.73$ & $77.00$ \\
SciBERT & $37.84$ & $81.46$ & $85.78$ & $88.42$ & $89.85$ & $62.21$ & $88.50$ & $90.94$ & $78.13$ \\
ManufactuBERT & $\mathbf{50.89}$ & $\mathbf{85.35}$ & $85.78$ & $\mathbf{92.66}$ & $\mathbf{91.49}$ & $68.11$ & $89.17$ & $91.28$ & $\mathbf{81.84}$\\ 
ManufactuBERT\textsubscript{D} & $49.52$ & $85.18$ & $\mathbf{87.25}$ & $91.40$ & $91.13$ & $\mathbf{68.59}$ & $\mathbf{89.80}$ & $\mathbf{91.36}$ &  $81.78$ \\
\bottomrule
\end{tabular}
\caption{\textbf{GLUE benchmark results}: We report accuracy for MNLI, MRPC, QNLI, QQP, RTE and SST-2, Matthew's correlation for CoLA and the mean of Pearson and Spearman correlations for STS-B. Results for RoBERTa are from \citet{liu2019robertarobustlyoptimizedbert}.}
\label{tab:glue_results}
\end{table*}

To verify the statistical significance of these improvements, we perform a pairwise model comparison using the Almost Stochastic Order (ASO) test \citep{del2018optimal, dror2019deep}, following the implementation by \citet{ulmer2022deep}. With a confidence level of $\alpha$=0.05 (adjusted with a Bonferroni correction) and $\tau$=0.5, the results confirm the superiority of ManufactuBERT\textsubscript{D}. It is stochastically dominant over its non-deduplicated version on 5 tasks and over the next best model, MatSciBERT, on 5 tasks. Its advantage is even more pronounced when compared to general-purpose models, where it dominates RoBERTa and NeoBERT on all nine tasks and demonstrates clear superiority over the remaining baselines on the majority of tasks. Almost stochastic dominance on each dataset will be provided in the Appendix.

Recent models NeoBERT and ModernBERT exhibit unexpectedly low performance on NER tasks. As these models were not originally evaluated on NER by their authors, the cause of this degradation remains unclear. Several open issues reported on GitHub at the time of writing suggest that similar behavior occurs on other benchmarks. Further analysis is needed to better understand this phenomenon.

\subsubsection{GLUE}

To quantify the impact of domain adaptation on general language understanding capabilities or \textit{catastrophic forgetting}, we evaluate our models on the GLUE benchmark. The primary goal is to measure the degree of performance degradation relative to our starting checkpoint, RoBERTa, and to compare this degradation against that of other specialized models, SciBERT and MatSciBERT.

The results, summarized in Table \ref{tab:glue_results}, indicate that while our models expectedly exhibit a performance drop compared to the general-domain RoBERTa baseline, they retain their foundational language abilities far more effectively than other specialized models. Both ManufactuBERT and ManufactuBERT\textsubscript{D} consistently and significantly outperform SciBERT and MatSciBERT on every GLUE task. On average, our models score over 4.5 points higher than MatSciBERT and 3.6 points higher than SciBERT on the same experimental setup.

This result shows that our adaptation recipe is effective. By continuing pretraining on a broad but domain-filtered web corpus, we preserve general language knowledge while adding manufacturing-specific expertise. In contrast, models like SciBERT, trained from scratch on academic text, tend to overfit to their domain and lose general understanding. Our approach achieves a better balance between domain knowledge and general language ability.

\section{Analysis}

\subsection{Data Deduplication Cost}

To quantify the impact of our data curation on training efficiency, we analyzed the convergence speed of ManufactuBERT (trained on the filtered corpus) and ManufactuBERT\textsubscript{D} (trained on the deduplicated corpus). We tracked their downstream performance on the FabNER dataset by evaluating checkpoints saved every 500 pretraining steps. Each evaluation was performed over 10 random seeds using the fine-tuning hyperparameters described in Section \ref{sec:hyperparameters_finetuning}.

\begin{figure}[!ht]
\begin{center}
\includegraphics[width=\columnwidth]{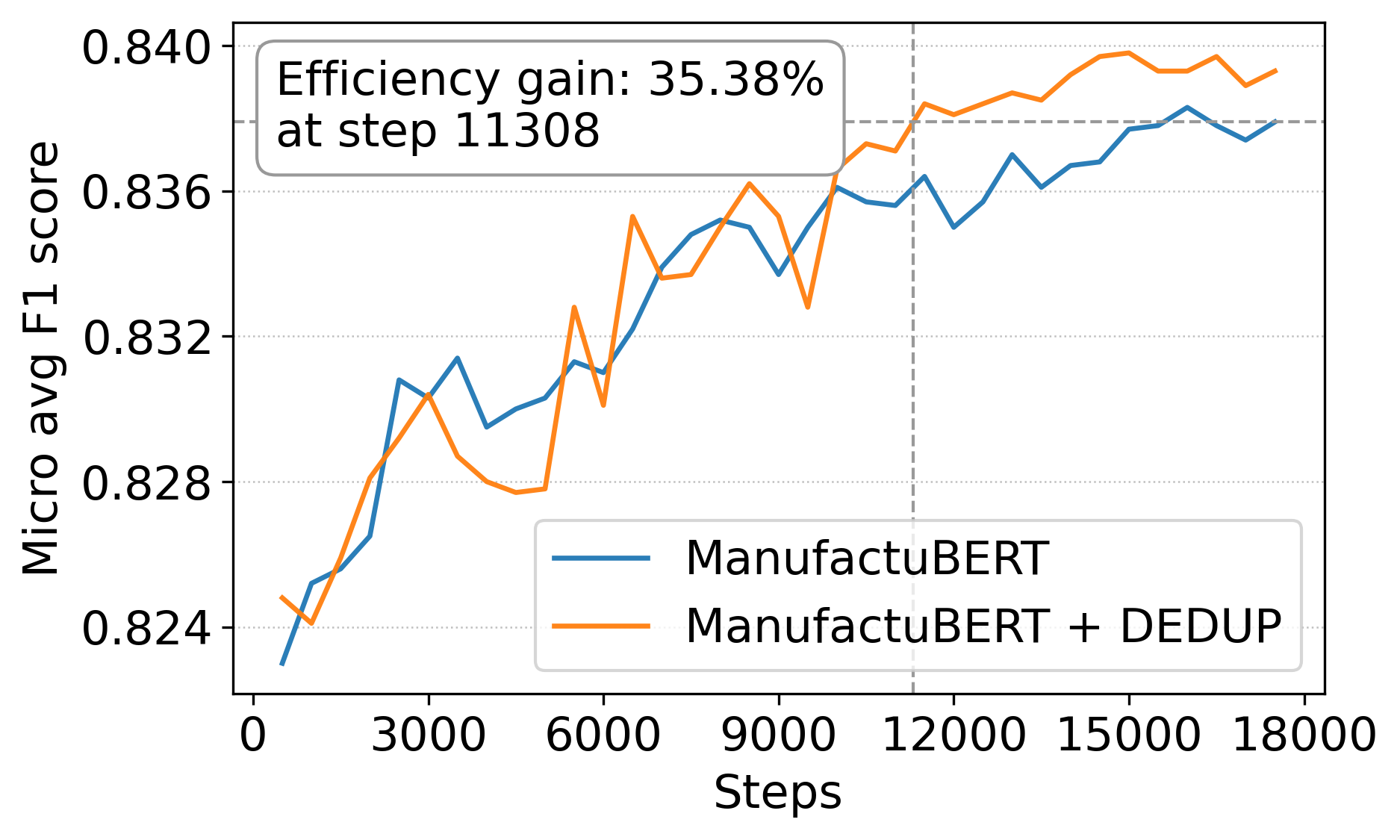}
\caption{Performance evolution of ManufactuBERT and ManufactuBERT\textsubscript{D} on the FabNER dataset across 17,500 training steps.}
\label{fig:efficiency}
\end{center}
\end{figure}

As illustrated in Figure \ref{fig:efficiency}, the model trained on the deduplicated corpus converges faster. We estimate that ManufactuBERT\textsubscript{D} achieves the final performance of the baseline ManufactuBERT at approximately step 11,308. This corresponds to a 35\% reduction in the required training iterations to reach the same performance level.

This accelerated convergence can be translated into energy savings. The full pretraining schedule of 17,500 steps on 8 NVIDIA V100 GPUs (250W TDP each) consumes an estimated 102,200 Wh. Reaching the equivalent performance at 11,308 steps would require only 66,032 Wh. However, this saving must be offset by the cost of the deduplication pipeline itself. As detailed in Table \ref{tab:dedup_cost}, the consumption  for this process can be estimated at 2,606.3 Wh. Taking this overhead into account, the net efficiency gain from employing deduplication is approximately 32.8\%.

\begin{table*}[!ht]
    \centering
    \begin{tabular}{lrr}
    \toprule
    \multicolumn{1}{l}{\textbf{Deduplication Stage}} &
    \multicolumn{1}{c}{\textbf{Time}} &
    \multicolumn{1}{c}{\textbf{Consumption (Wh)}} \\ \midrule
     MinHash Signatures & 35h x 8 CPU & 1 050 \\
     MinHash Buckets & 1.4h x 8 CPU & 42 \\
     MinHash Clusters & 0.21h x 8 CPU & 6,3 \\
     MinHash Filtering & 0.33h x 8 CPU & 10 \\ 
     MinHash Total & & 1108,3 \\ \midrule
     SemDeDup Tokenizer & 0.8h x 8 CPU & 24 \\
     SemDeDup Embeddings & 3.6h x 1 A100 & 1 440 \\
     SemDeDup Clustering & 0.16h x 32 CPU & 19,2 \\
     SemDeDup Deduplication & 0.16h x 8 CPU & 4,8 \\
     SemDeDup Filtering & 0.33h x 8 CPU & 10 \\ 
     SemDeDup Total & & 1498 \\ \midrule
     \textbf{Total} & & 2606.3 \\ \bottomrule
     
    \end{tabular}
    \caption{Computational cost and energy consumption of the deduplication pipeline. CPU consumption is based on Intel Cascade Lake 6248 cores (3.75W TDP per core), and GPU consumption is based on NVIDIA A100s (400W TDP).}
    \label{tab:dedup_cost}
\end{table*}

\subsection{Comparison of Training Costs}

We compare our computational costs against those of related domain-specific models. The authors of MatSciBERT report a training time of 15 days on two V100 GPUs to adapt their model from a SciBERT checkpoint, which equates to approximately 720 V100-hours. The pretraining of SciBERT itself, conducted from scratch, required one week on a TPUv3-8. Given that this hardware is comparable to a four-V100 compute cluster\footnote{\url{https://timdettmers.com/2018/10/17/tpus-vs-gpus-for-transformers-bert/}}, this training phase represents an estimated 672 V100-hours.

In contrast, our adaptation of RoBERTa required only 408 V100-hours to complete the 17,500 steps (51 hours on 8 V100s). Therefore, our methodology is not only more performant on downstream tasks but is also more computationally efficient than other specialized models in the literature.

\subsection{Effect of Deduplication Granularity}

\begin{figure}[!ht]
\begin{center}
\includegraphics[width=\columnwidth]{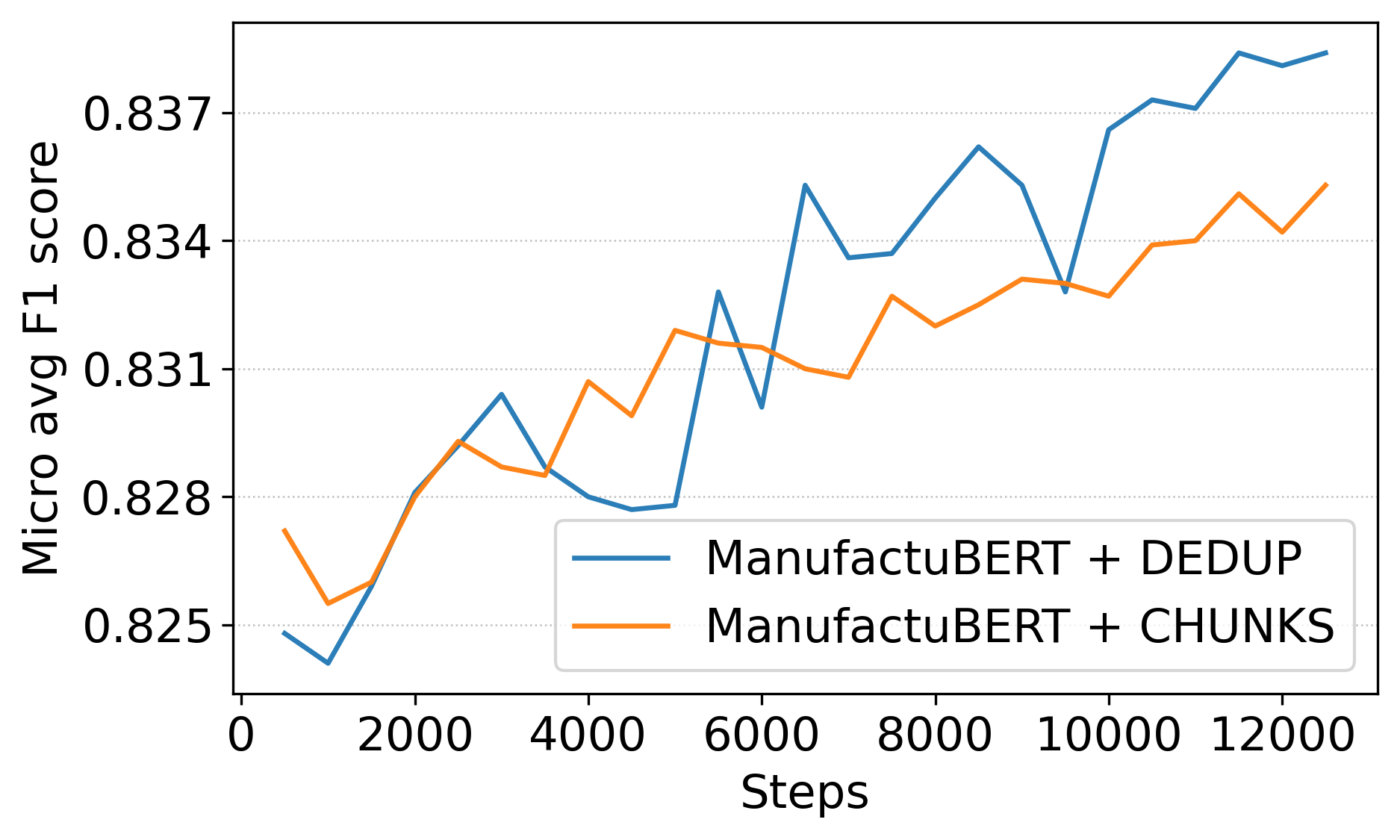}
\caption{Performance evolution of ManufactuBERT\textsubscript{C} and ManufactuBERT\textsubscript{D} on the FabNER dataset across 12,500 training steps.}
\label{fig:chunk_semdedup}
\end{center}
\end{figure}

In our original SemDeDup setup, each document was represented by the average embedding of its 512-token chunks. 
To determine if this aggregation strategy was optimal, we investigated a more fine-grained alternative by applying SemDeDup at the chunk level. In this setup, we deduplicated the set of all 512-token chunks across the entire corpus before pretraining. We then trained a new model, ManufactuBERT\textsubscript{C}, on this chunk-deduplicated dataset. We evaluated this alternative model on the FabNER benchmark. 
As shown in Figure~\ref{fig:chunk_semdedup}, this more granular approach does not yield any improvement, despite requiring more computation due to the larger number of clustering points. To save computational resources, we restricted this comparative experiment to the first 12,500 pretraining steps.

\subsection{Further Deduplication Using D4}

\begin{figure}[!ht]
\begin{center}
\includegraphics[width=\columnwidth]{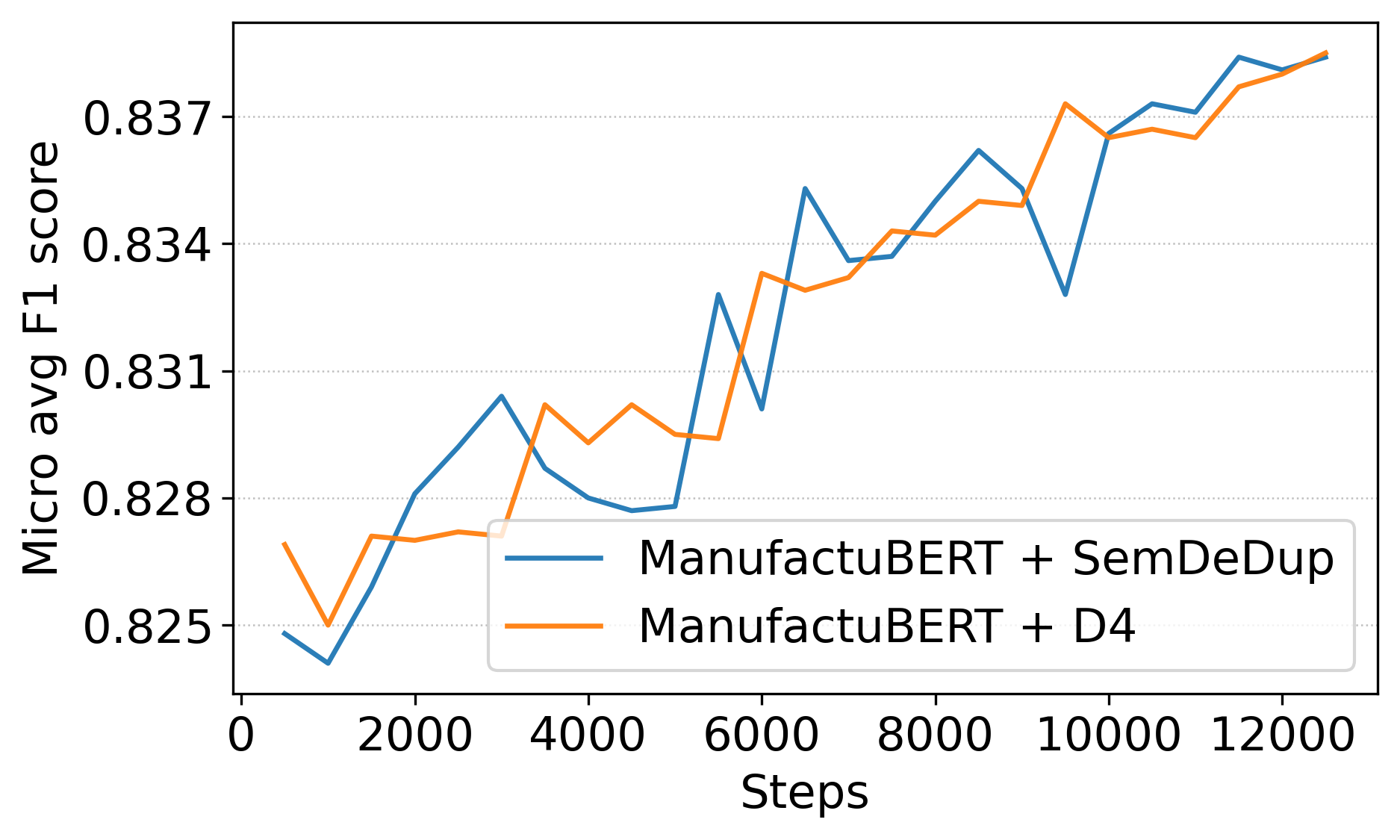}
\caption{Performance evolution of ManufactuBERT\textsubscript{D4} and ManufactuBERT\textsubscript{D} on the FabNER dataset across 12,500 training steps.}
\label{fig:D4_results}
\end{center}
\end{figure}

We also investigated whether more aggressive data pruning could further improve performance. D4 \cite{tirumala2023d4improvingllmpretraining} is a refinement of the SemDeDup algorithm. The authors observed that some clusters in SemDeDup are highly redundant, and that dense regions of the embedding space can degrade the quality of K-means clustering. To address this, D4 introduces two additional steps after the SemDeDup algorithm:

(1) reclustering the already deduplicated dataset

(2) retaining only the points farther to each cluster centroid according to a ratio, denoted as \textit{R\textsubscript{proto}}.

We applied D4 to our deduplicated dataset using \textit{R\textsubscript{proto}} = 0.75 and pretrained a new model, ManufactuBERT\textsubscript{D4}. We then evaluated this model on the FabNER benchmark. As shown in Figure \ref{fig:D4_results}, this more aggressive deduplication strategy did not yield any additional performance gains. Moreover, D4 introduces two extra deduplication stages, increasing the overall computational cost. We also limited this comparative experiment to the first 12,500 pretraining steps.

\section{Conclusion}

In this work, we introduce ManufactuBERT, a RoBERTa-based language model continually pretrained on a large-scale corpus specifically curated for the manufacturing domain. By filtering the FineWeb corpus and then applying deduplication, we construct a compact, high-quality pretraining corpus.

Our empirical results validate this approach : ManufactuBERT achieves new state-of-the-art results on several manufacturing-related NLP benchmarks while maintaining strong general-domain capabilities on GLUE.

Beyond its results, our work offers a reproducible and efficient framework for domain adaptation of encoder-based language models and underscores the importance of data quality and diversity over dataset size.

Future work will explore extensions of this pipeline to other specialized domains, integration with modern encoder architectures, and alternative data selection and deduplication algorithms.

\section{Limitations}

Despite its strong performance, ManufactuBERT faces several limitations:

Our pretraining dataset is derived from web data and may not accurately represent documents encountered in real manufacturing contexts, which are often internal or confidential.

Our adaptation currently targets only English texts, limiting the model's applicability to multilingual or non-English manufacturing data.

Finally, although the deduplication pipeline improves data efficiency, it introduces additional preprocessing overhead, which may limit scalability for larger corpora or yield reduced benefits for smaller datasets or low-resource domains.

\section*{Acknowledgments}

This publication was made possible by the use of the FactoryIA supercomputer, financially supported by the Ile-De-France Regional Council. It also benefited from the support of the DataFIX project, financed by the French government under the France 2030 Programme and operated by Bpifrance.

\nocite{*}
\section{Bibliographical References}\label{sec:reference}

\bibliographystyle{lrec2026-natbib}
\bibliography{lrec2026-example}


\end{document}